\pdfoutput=1

\documentclass[11pt]{article}

\usepackage[final]{acl}
\usepackage{multirow}
\usepackage{times}
\usepackage{latexsym}
\usepackage{algorithm}
\usepackage{algorithmic}
\usepackage{graphicx}
\usepackage{booktabs}
\usepackage{amsmath}
\usepackage{makecell}
\usepackage{enumitem}
\usepackage[utf8]{inputenc} 
\usepackage[T1]{fontenc}    
\usepackage{hyperref}       
\usepackage{url}            
\usepackage{booktabs}       
\usepackage{amsfonts}       
\usepackage{nicefrac}       
\usepackage{microtype}      
\usepackage{xcolor}         
\usepackage{mdframed}
\usepackage{threeparttable}
\usepackage{graphics}
\usepackage{multirow}
\usepackage{rotating}
\usepackage{fontsize}
\usepackage{array}
\usepackage{caption}
\usepackage{xcolor}
\usepackage{graphicx}
\usepackage{colortbl}
\usepackage{listings}
\usepackage{enumitem}
\usepackage{amsmath}
\usepackage{amssymb}
\usepackage{cleveref}
\usepackage{hyperref}
\usepackage{xspace}
\usepackage[T1]{fontenc}

\usepackage[utf8]{inputenc}

\usepackage{microtype}

\usepackage{inconsolata}

\usepackage{graphicx}
\usepackage{tabularx}
\usepackage[colorinlistoftodos]{todonotes}

\title{JAM: Controllable and Responsible Text Generation via Causal Reasoning and Latent Vector Manipulation}

\author{Yingbing Hunag, Deming Chen, \and Abhishek K. Umrawal \\
        University of Illinois Urbana-Champaign \\ \texttt{\{yh21, dchen, aumrawal\}@illinois.edu}}

\begin{document}
\maketitle
\begin{abstract}
While large language models (LLMs) have made significant strides in generating coherent and contextually relevant text, they often function as opaque black boxes, trained on vast unlabeled datasets with statistical objectives, lacking an interpretable framework for responsible control. In this paper, we introduce JAM (Just A Move), a novel framework that interprets and controls text generation by integrating cause-effect analysis within the latent space of LLMs. Based on our observations, we uncover the inherent causality in LLM generation, which is critical for producing responsible and realistic outputs. Moreover, we explore latent vectors as fundamental components in LLM architectures, aiming to understand and manipulate them for more effective and efficient controllable text generation. We evaluate our framework using a range of tools, including the HHH criteria, toxicity reduction benchmarks, and GPT-4 alignment measures. Our results show that JAM achieves up to a 22\% improvement over previous Controllable Text Generation (CTG) methods across multiple quantitative metrics and human-centric evaluations. Furthermore, JAM demonstrates greater computational efficiency compared to other CTG methods. These results highlight the effectiveness and efficiency of JAM for responsible and realistic text generation, paving the way for more interpretable and controllable models.
\end{abstract}

\begin{figure*}[t]
        \centering
        \includegraphics[width=\textwidth]{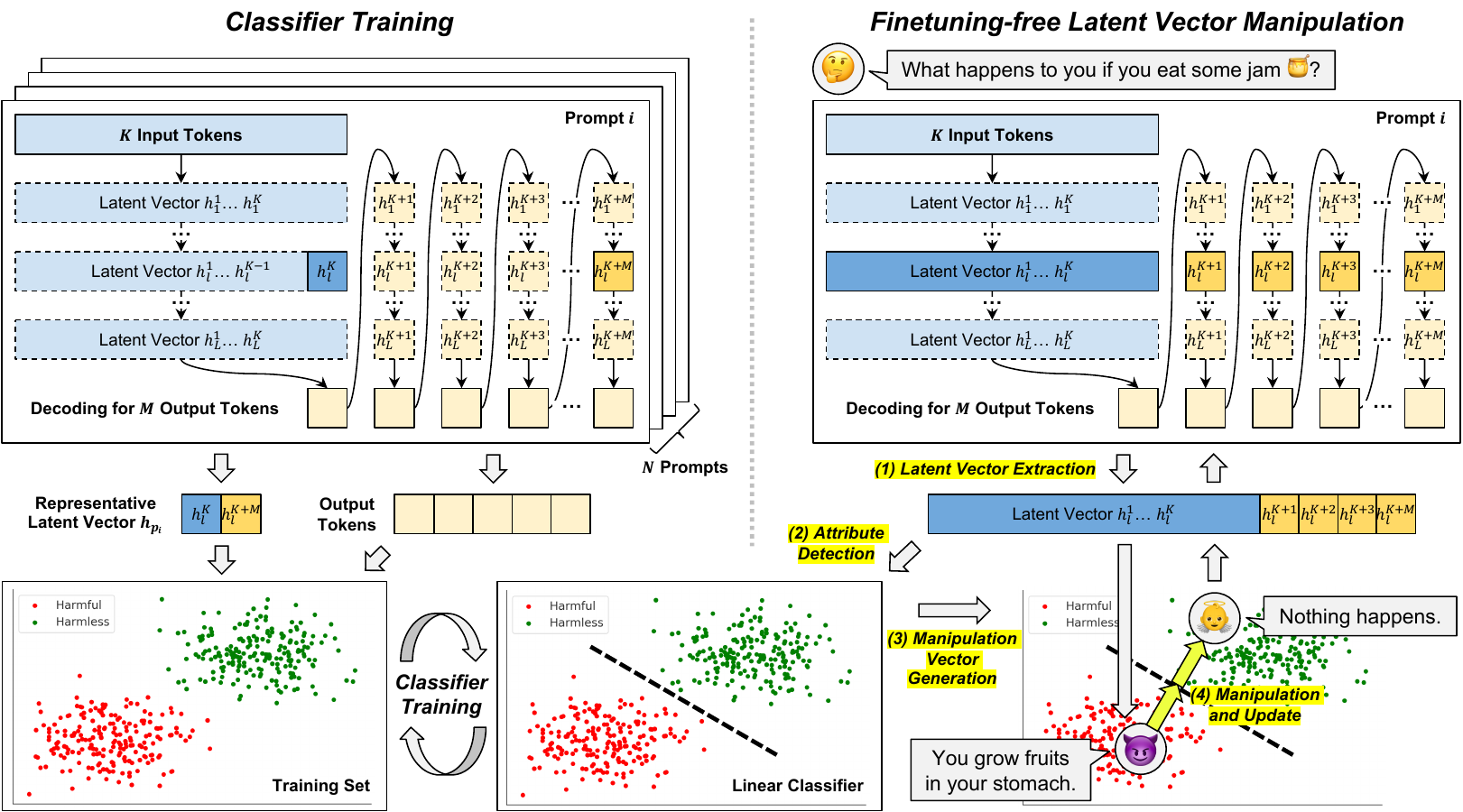}
        \caption{The framework of JAM. We first train a binary linear classifier regarding to an attribute, such as harmless in scenarios. The details for classifier training are in Sec. \ref{sec:classifier}. During inference, it contains four steps: latent vector extraction, attribute detection, manipulation vector generation, and manipulate and update generation. The details for the inference are in Sec. \ref{sec:framework}.}
        \label{fig: jam_flow}
\end{figure*}

\section{Introduction}
Large Language Models (LLMs) have recently demonstrated remarkable capabilities across various domains, including question-answering, content creation, and reasoning. Chatbots, as one of the most popular interfaces for LLMs, engage diverse audiences—ranging from children to corporations—and these different user groups often prefer content with specific attributes. Given the expansive user base and various application scenarios, it is increasingly important to generate controllable content that aligns with desired attributes.

\noindent \textbf{Controllable Text Generation (CTG)} methods have been more and more studied recently; one of the most common techniques for controlling LLMs is prompting \cite{NEURIPS2022_b1efde53}. While prompting is straightforward to implement, it often falls short in steering LLMs in specific directions and producing reliable outputs consistently. Moreover, prompting cannot help people understand and interpret the generation process of LLMs, which is crucial for exerting precise control over their outputs. Apart from prompting-based methods, some works rely on training or fine-tuning models to help control generation \cite{yang2021fudge, meng2022controllable, li2024reinforcement, alhafni2024personalized}, which by nature get approximate solutions and do not guarantee that the desired constraints are satisfied. This method typically demands intensive computational resources and poses challenges in scaling up one fine-tuned model to accommodate the different desired attributes. For test- or inference-time methods, Mix and Match \cite{mireshghallah2022mixmatchlearningfreecontrollable} and FreeCtrl \cite{feng2024freectrl} continuously evaluate an attribute during generation, which could bring extra overhead to decoding steps.

\noindent \textbf{Causal reasoning} offers a valuable framework for investigating and understanding the generative processes of LLMs \cite{xu2021causalitylearningnewperspective,leclair2021ensemblemodelsneuralsource, kaddour2022causalmachinelearningsurvey}. Traditional methods for controlling LLM outputs, such as prompting or fine-tuning, often treat the models as black boxes, making it difficult to predict or interpret their behaviors comprehensively. By contrast, causal reasoning enables us to model the internal mechanisms of LLMs more transparently, identifying how different components and latent variables influence the generated text. We believe that incorporating causality is essential for enhancing controlled text generation because it allows for systematic intervention in the generative process to achieve desired outcomes.

\noindent \textbf{Latent spaces in LLMs} are outputs of intermediate layers during computation, and serve as vital latent representations that encode the information and understanding LLMs utilize during text generation. These latent spaces convey various aspects of language, such as semantics, logics \cite{chen2024stateshiddenhiddenstates}, and hallucinations \cite{duan2024llms}, and are crucial for the model's ability to generate coherent and contextually appropriate responses. Therefore, by exploring and manipulating these latent vectors through the lens of causality, we see the potential to identify the cause-and-effect relationships that govern the model's behaviors. This approach should not only provide deeper insights into how LLMs process information but also establish a rigorous method to steer them toward generating reliable and desired responses.

In this paper, we investigate the causality inherent in popular LLMs and examine the consistency of inferred causal directions in Helpful, Honest, and Harmless (HHH) criteria tasks \cite{askell2021general} and other tasks, e.g., toxicity reduction \cite{gehman2020realtoxicityprompts}. Based on our observations of causal reasoning and latent vectors during generation, we propose a new framework, \textbf{JAM}~ (Just A Move) as shown in Figure \ref{fig: jam_flow}, that could control LLM generation by just a small move on latent vectors while preserving the original models' causality. JAM also ensures the quality and reliability of the generated outputs. For example, by intervening on specific latent states associated with undesirable attributes, we can suppress or modify these attributes in the output, thereby achieving more controlled and aligned text generation. Our contributions can be summarized as follows: 
\begin{enumerate}
    \item We investigate the behavior of latent vectors during LLM generation and uncover the existence of causal relationships within the latent spaces, specifically for attributes in alignment tasks. We also provide statistical evidence for more insights. (Sec. \ref{sec:latent})
    \item Based on our observation, we propose the JAM framework shown in Figure \ref{fig: jam_flow}, which includes attribute classifier training and latent vector manipulation to control LLM generation with negligible overhead. (Sec. \ref{sec:method})
    \item We conduct experiments on HHH-criteria and toxicity reduction on popular LLMs including Mistral-7B and Llama3-8B. JAM can improve up to 10\% scores on multiple metrics and win most of the time on GPT-4 evaluation compared with the traditional method. (Sec. \ref{sec:exp})
    \item We conduct ablation studies on the choices of parameters to provide a more robust analysis and potential interpretations of the behaviors of LLMs' latent vectors. (Sec. \ref{sec:exp})
\end{enumerate}
\section{Background and Observations}
\label{sec:latent}
\subsection{Latent Space in LLMs}
\label{sec:gene}
Latent spaces have been investigated and used in many previous works on LLMs. The work \citet{bronzini2024unveiling} explores how factual knowledge is encoded into the latent space of different layers and utilizes this observation to facilitate tasks such as fact-checking and semantic understanding. Latent vectors are also considered to represent the intentions for LLMs generation \cite{jiang2023latent}. All these previous works demonstrate the importance and potential of understanding and using latent vectors for the controllable generation of LLMs. However, none of the previous research has comprehensively explored the causality within latent spaces and its effects on LLM generation. To further investigate how latent vectors play a role during generation, we use HH-RLHF data from \citet{bai2022training}, the Honest dataset from \citet{lin2022truthfulqameasuringmodelsmimic} and conduct the following experiments on \texttt{Meta-Llama-3-8B} \cite{llama3modelcard}:

\begin{figure}[t]
    \centering
    \includegraphics[width=0.5\textwidth]{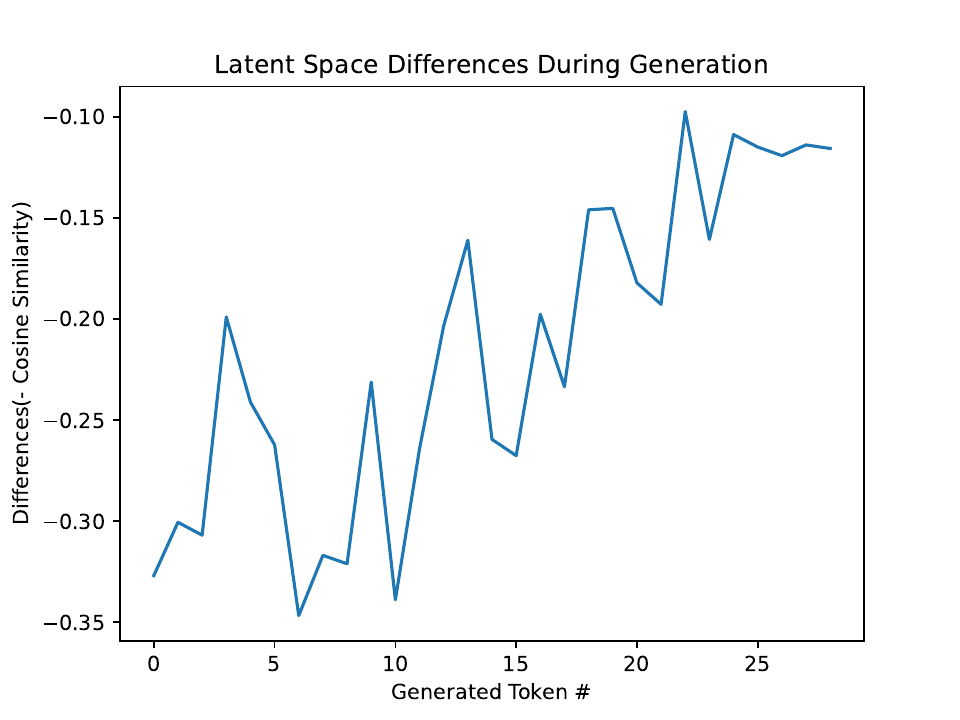}
    \caption{The average differences among the latent vectors of LLMs at each decoding step across all layers. }
    \label{fig: gene}
\end{figure}

\begin{figure}[t]
        \centering
        \includegraphics[width=0.5\textwidth]{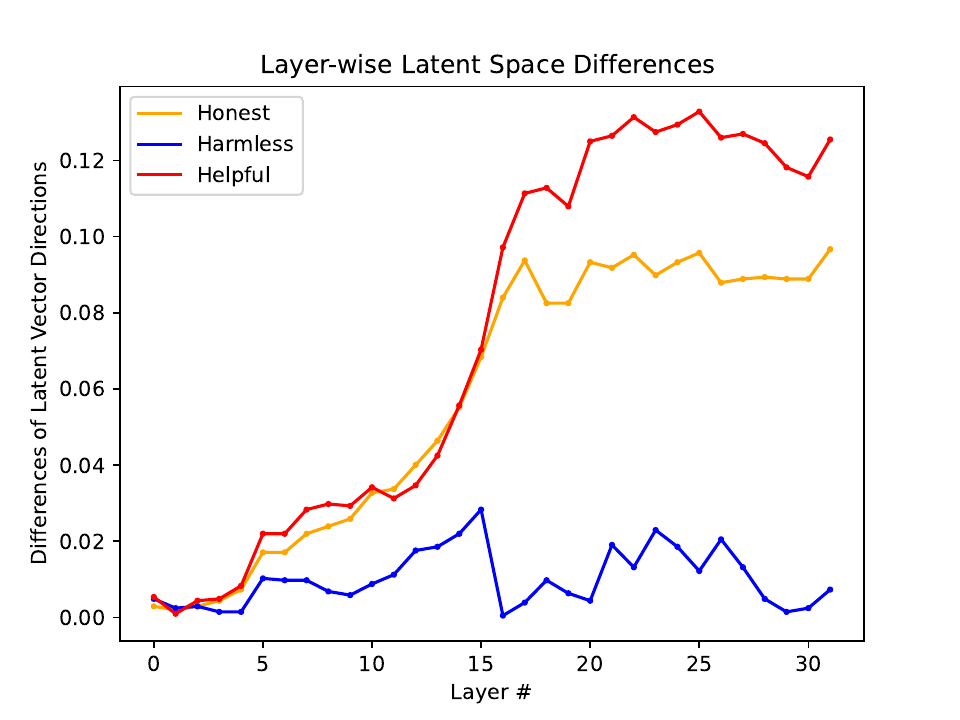}
        \caption{The average differences among latent vectors of LLMs for each layer. Different colors represent different tasks. Harmless and helpful datasets are from HH-RLHF \cite{bai2022training} and honest dataset from TruthfulQA \cite{lin2022truthfulqameasuringmodelsmimic}.}
        \label{fig: layer}
\end{figure}

\begin{itemize}[topsep=0pt,parsep=4pt,itemsep=0pt,leftmargin=15pt]
    \item  \textbf{The evolution and pattern of latent vectors during generation.}
    To understand how latent vectors develop and enable LLMs to output answers corresponding to distinct prompts, it is necessary to investigate the differences between latent vectors along different decoding steps. We select $N$ prompts $\{p_1 ... p_N\}$ each with $K$ tokens and let LLMs generate $M$ tokens to each one. We save the latent vectors, marked as $h^m_{i|l}$, of the generated token at layer $l$ of decoding step $m$ for prompt $p_i$. Then, we calculate the differences among $\{h^m_{1|l}...h^m_{N|l}\}$ as negative cosine similarity: 
    \begin{equation}\label{eq: cos_sim}
    \small
        \mathbf{d}(h^m_{i|l}, h^m_{j|l}) = - \frac{h^m_{i|l} \cdot h^m_{j|l}}{\|h^m_{i|l}\| \|h^m_{j|l}\|}
    \end{equation}
    
    where $i,j \in [1,N], m \in [1, M], l \in [1,L]$. In this experiment, we set $N$ to 10 and $M$ to 30, and compute average $\mathbf{d}(h^m_{p_i|l}, h^m_{p_j|l})$ for each decoding layer over $N$ prompts. As illustrated in Figure \ref{fig: gene}, we observe that the latent vectors progressively diverge as the generation process advances, indicating that they become increasingly specific and representative at later decoding steps. 
    
    \item \textbf{The layer-wise behaviors of latent vectors.} 
    \label{sec:layer}
    It is also crucial to analyze layer-wise behaviors to gain deeper insights into LLM generation. From the previous experiments, we observe that the latent vector of the last token during generation $h^{K+M}_i$ is more expressive, and the latent vector of the last token in the prefilling process $h^K_i$ encapsulates previous tokens' information of the prompt according to attention mechanism. To simplify while preserving essential information for each prompt $p_i$, we extract the latent vectors from these two tokens for every layer $l$ and concatenate them to form a new latent vector $h_{p_i|l}$. 
    \begin{equation}
    \label{eq:latent}
        h_{p_i|l}=[h^K_{i|l}, h^{K+M}_{i|l}]
    \end{equation}

    We calculate $\mathbf{d}(h_{p_i|l}, h_{p_j|l})$ according to Eq. (\ref{eq: cos_sim}) for every layer $l$. The results in Figure \ref{fig: layer} demonstrate that the latent vectors become increasingly stable and representative in layers beyond $15^{th}$, conveying more specific information. This characteristic allows for more effective control over LLM generation, enabling more precise manipulation of the output.
    
\end{itemize}

\subsection{Causality in LLMs}
\label{sec:causality}
Causal reasoning has been proven to be effective for generative tasks in different areas \cite{bose2022controllable, geiger2022inducing, stolfo2022causal}, and it enables generative models to produce more realistic and reliable output. However, the sequential nature of the language models makes it significantly different and more complex to achieve controllable text generation compared to other models. In this section, we aim to reveal the existence of causality in latent spaces of LLM generations and the importance of considering causality in CTG. There are two research questions we want to answer:
\begin{enumerate}
    \item \textbf{If each attribute in LLM generation is separable by a linear binary classifier.} Such separability would suggest that the underlying structure of the latent space is well-organized and distinct for different attributes, making it possible to detect and manipulate specific attributes with minimal complexity and high efficiency, facilitating attribute-based control over the generated content.
    \item \textbf{If we can identify causality existing in LLM generation.} Understanding these causal links is essential for controlling the generation process in a way that retains the natural flow of language and ensures that interventions (such as manipulating certain attributes) result in coherent and logically consistent text. This could lead to more reliable, ethical, and explainable content.
\end{enumerate}

\begin{table}[t]
    \centering
    \caption{Latent vectors SVM classifier test accuracy.}
    \label{table:svm}
    \begin{tabular}{lcc}
        \toprule
        \textbf{Attributes} & \textbf{SVM Classifier Accuracy} & \textbf{F1} \\ \midrule
        Honest & 0.80 & 0.78 \\ 
        Helpful & 0.72 & 0.71 \\ 
        Harmless & 0.71 & 0.70\\ \bottomrule
    \end{tabular}
\end{table} 

\noindent \textbf{Classifier.} 
Binary classifiers that predict the likelihood of an attribute based on input text can be employed to shape the output distribution to align with the specified attribute \cite{sitdikov2022classifiersbetterexpertscontrollable, liang2024controlled, dekoninck2024controlledtextgenerationlanguage}. 
These classifiers can capture attributes that are not easily detectable in natural language. Unlike previous approaches that use text as input, we utilize latent vectors, aiming to explain attributes at a more fundamental level. Additionally, the linear model offers more interpretability compared to more complex models, providing further opportunities to understand and leverage the linear classifier effectively.

In our experiments, we train a linear binary classifier with decision boundary as hyperplane $S_A$ for latent vectors $h_{pi|l}$, as defined in Eq. (\ref{eq:latent}). This is done for each attribute in the HHH criteria using datasets containing over 1000 data points per attribute, with an 8:2 train-test split. Based on our observation in Figure \ref{sec:latent}, layers beyond the $20^{th}$ are more expressive. Thus, for all subsequent experiments, we set $l$ to $20^{th}$ layer for effectiveness, and simplify $h_{p_i|l}$ to $h_{p_i}$. The test accuracy results of Support Vector Machines (SVM) in Table \ref{table:svm} suggest that attribute patterns can be effectively captured in a linear manner.

\begin{table}[t]
    \centering
    \caption{Cause-effect correlation analysis with correlation coefficient ($\rho$) and $p$ value ($p$) on HHH criteria for both directions.}
    \label{table:causality}
    \begin{tabular}{llcc}
        \toprule
        \textbf{Cause} & \textbf{Effect} & \textbf{$\rho$} & \textbf{$p$} \\
        \midrule
        Honest & Harmless & 0.23 & 1.37e-024 \\ 
        Honest & Helpful & 0.66 & 2.50e-250 \\ 
        Harmless & Honest & 0.61 & 4.11e-149 \\ 
        Harmless & Helpful & 0.59 & 8.37e-140 \\ 
        Helpful & Honest & 0.63 & 1.20e-214 \\ 
        Helpful & Harmless & 0.24 & 9.34e-027 \\ \bottomrule
    \end{tabular}
\end{table}

\noindent \textbf{The existence of causality.} Consider a latent vector $h_{p_i}$, and a cause attribute $A_c$, where $A_c=1$ means this attribute exists in answers and 0 means it doesn't exist. Hyperplane $S_{A_c}$ is the decision boundary of the classifier trained for detecting the cause attribute $A_c$. The hidden state is controlled on $A_c$ by: 
\begin{align}
\label{eq:cf_mani}
    T_{A_c}(h, \alpha) &= h + \alpha S_{A_c} \\
    h_{p_i|A_c=1} &= T_{A_c}(h_{p_i|A_c=0} , \alpha)
\end{align}
$T_{A_c}$ is the latent vector manipulation function $R^n \times R \to R^n$ that shifts the latent vector $h_{p_i|A_c=0}$ across the decision hyperplane by a distance $\alpha$. The value of $\alpha$ is determined analytically as the minimum displacement required to move $h_{p_i|A_c=0}$ across the decision boundary, thereby flipping its classified label from 0 to 1. By applying $T_{A_c}$ , the latent vector $h_{p_i|A_c=0}$  is transformed into  $h_{p_i|A_c=1}$ , ensuring that the classifier assigns it a label of 1. The reverse manipulation, shifting  $h_{p_i|A_c=1}$  back to  $h_{p_i|A_c=0}$, follows the similar principle

Given latent vector $h_{p_i|A_c=0}$, we calculate its causality effect $\phi$ on an effect attribute $A_e$ as:
\begin{align}
\label{eq:effect}
    \phi = |S_{A_e} \cdot T_{A_c}(h_{p_i|A_c=0} , \alpha) - S_{A_e} \cdot h_{p_i|A_c=0}|
\end{align}
Hyperplane $S_{A_e}$ serves as the decision boundary for the classifier trained to detect the effect attribute $A_e$. The classifier predicts the label by computing $S_{A_c} \cdot h_{p_i|A_c=1}$. If $\phi=1$, it means the change of the label for cause attribute $A_c$ also causes the label for effect attribute $A_e$ to flip, $\phi=0$ otherwise. A similar logic is also followed by given input $h_{p_i|A_c=1}$. In Table \ref{table:causality}, we provide the causal effect estimates and their statistical significance for both directions. Here, all $p <= 0.05$ are typically considered statistically significant, and it indicates confidence in the existence of a causal relationship.

It is important to explore and consider the causal relationships among attributes. When aiming to control a specific attribute, preserving the existing causality in LLMs is key to ensuring that the controlled text generation remains reliable and more aligned with natural language. We present a quantitative analysis of the effectiveness of this causal approach in Sec. \ref{sec:exp_gpt}, where the GPT-4 model demonstrates a preference for answers that account for causal relationships.
\section{JAM Framework}
\label{sec:method}
With the observations summarized in Sec.~\ref{sec:latent}, we propose the JAM framework shown in Figure~\ref{fig: jam_flow} to train the attribute classifier and control the LLM output through latent vector manipulation.

\subsection{Data Collection and Classifier Training}
\label{sec:classifier}
For the target LLMs, the dataset consists of prompts, LLM's answers, and the ground truth label given by the annotator regarding an attribute $A$. To automate the process in this project, we use pairs of correct and incorrect answers for each prompt instead of relying on the given ground truth. Correct answers convey the desired attribute, while incorrect answers do not. We use a text-to-text transfer transformer (T5) \cite{2020t5} to assign the label. It compares each generated answer with a correct and incorrect examples pair, assigning a label $y$ of 1 if the answer is closer to the correct examples and vice versa. After extracting latent vector $h_{p_i}$ described by Eq. (\ref{eq:latent}), we create dataset as \{$h_{p_i}, y_i$\}. The data is split into a training and testing set with an 8:2 ratio, totaling 500 data points, to train our binary classifier, as illustrated in the bottom left of Figure \ref{fig: jam_flow}. For classifier selection, we explore both SVM and logistic regression models, finding no significant difference in their performance.

\begin{algorithm}
\caption{Pseudo Code of JAM Algorithm}
\label{algo1}
\begin{algorithmic}[1]
    \STATE \textbf{Input:} Prompt dataset \{$p_1...p_N$\} with size $N$. LLM model $M$, and desired attribute $A$.
    \STATE \textbf{Output:} Answers \{$a_1...a_N$\} with desired attributes.
    \FOR{$p_i$ in \{$p_1...p_N$\}}
        \STATE \textcolor{gray}{\textit{\# extract latent vectors}}
        \STATE $h^1_i ... h^{K+M}_i \gets M(p_i)$
        
        \STATE $h_{p_i} \gets [h^K_i, h^{K+M}_i]$
        \STATE \textcolor{gray}{\textit{\# detect attribute on attribute $A$}}
        \STATE $label \gets S_A.T \cdot h_{p_i}$
        \STATE \textcolor{gray}{\# \textit{$label$ is 0: the desired attribute $A$ not detected}}
        \IF{$label == 0$}
            \STATE \textcolor{gray}{\textit{\# generate manipulation vector}}
            \STATE $\alpha \gets distance(h_{p_i},S_A)$
            \STATE \textcolor{gray}{\textit{\# manipulate and update generation}}
            \STATE Update $M(p_i)$ with  $\alpha S_A$
        \ENDIF \\
        $a_i = M(p_i)$
    \ENDFOR
    \STATE \textbf{Return} ${a_1...a_N}$
\end{algorithmic}
\end{algorithm}

\subsection{Latent Vector Manipulation During Inference} 

\label{sec:framework}
As shown in Figure~\ref{fig: jam_flow}, after classifier training, the JAM framework employs the classifier to manipulate latent vectors during inference. The manipulation process can be summarized into the following steps in Algorithm \ref{algo1}:
\begin{itemize}[topsep=0pt,parsep=4pt,itemsep=0pt,leftmargin=15pt]
    \item \textbf{Latent Vector Extraction:} we extract the representative latent vector for attribute detection and manipulation based on Eq. (\ref{eq:latent}). According to our observation in Sec. \ref{sec:latent}, this extracted vector conveys critical information for both the prefilling and generation stages, which aids the subsequent processes.
    \item \textbf{Attribute Detection:} during inference, given an input prompt and original answer, we first extract the latent vector $h_{p_i}$, and then let the trained classifier $S_A$ to decide if the answer contains the desired attribute $A$. If not, we move further to latent vector manipulation; otherwise, output the original answer.
    \item \textbf{Manipulation Vector Generation:} then we calculate the smallest distance $\alpha$ based on Eq.~(\ref{eq:cf_mani}) that could move $T_{A_c}(h_{p_i|A_c=0} , \alpha)$ to successfully generate vector $h_{p_i|A_c=1}$ with label 1. We use $\alpha S_A$ as a manipulation vector. 
    \item \textbf{Manipulate and Update Generation:} we update the latent vector of the last token of prefilling $h^K_{p_i}$ by adding the first half of the manipulation vector and gradually update the generated token by adding the second half of manipulation vector during each generation step.
\end{itemize}
\begin{table*}
    \setlength\tabcolsep{3pt}
    \centering
	\small
	\caption{Rouge2 and Bleurt scores on HHH-criteria tasks and perplexity of JAM with various moving distance (marked as model\_name w/ $\alpha$, i.e., \texttt{Llama3} w/ 1.2$\alpha^*$) compared with original \texttt{Meta-Llama-3-8B} and \texttt{Mistral-7B-Instruct-v0.2} models and their variants enhanced with previous CTG work, PREADD \cite{pei2023preadd}. Harmless and helpful datasets are from HH-RLHF \cite{bai2022training} and honest dataset from TruthfulQA \cite{lin2022truthfulqameasuringmodelsmimic}. Rouge2 and Bleurt scores are higher the better, perplexity is lower the better. We use bolded text to emphasize the model that achieves the best performance for each metric. The results from PREADD marked with~$^\text{x}$~are indicated as invalid because of the generated answers start being unreadable in the middle of the generation and largely affect the evaluation metrics.}
	\label{table:hhh}
	    \small
        \begin{tabularx}{\textwidth}{@{}
                l
                X  
                X  
                X
                X
                X
                X
                X
                X
                X
                X
                @{}
            }
            \toprule
            & \multicolumn{3}{c}{Harmless} & \multicolumn{3}{c}{Honest} & \multicolumn{3}{c}{Helpful} \\
            \cmidrule(lr){2-4}\cmidrule(lr){5-7}\cmidrule(lr){8-10}
             & \makecell{Rouge2}($\uparrow$) & {Bleurt}($\uparrow$) & {Perpl.}($\downarrow$) & \makecell{Rouge2}($\uparrow$)  & {Bleurt}($\uparrow$)  & {Perpl.}($\downarrow$)  & \makecell{Rouge2}($\uparrow$)   & {Bleurt}($\uparrow$)  & {Perpl.}($\downarrow$) \\
            \midrule
            \texttt{Llama3} & 0.205 & 0.485 & 37.73 & 0.227 & 0.441 & \textbf{3.60} & 0.230 & 0.580 & \textbf{14.43}\\
            \texttt{Llama3} w/ $\alpha^*$ & \textbf{0.250} & \textbf{0.530} & 36.13 & \textbf{0.318} & \textbf{0.463} & 3.72 & \textbf{0.300} & 0.550 & 17.11\\
            \texttt{Llama3} w/ 1.2$\alpha^*$ & 0.245 & 0.525 & \textbf{36.10} & 0.290 & 0.441 & 3.80 & 0.250 & \textbf{0.650} & 19.70 \\
            \texttt{Llama3} w/ 1.5$\alpha^*$ & 0.245 & \textbf{0.530} & 36.25 & 0.304 & 0.450 & 4.00 & 0.250 & 0.600 & 23.80 \\
            \texttt{Llama3} w/ \texttt{PREADD} & 0.315$^\text{x}$ & 0.485$^\text{x}$ & 171.10$^\text{x}$ & 0.245$^\text{x}$ & 0.545$^\text{x}$ & 177.16$^\text{x}$ & 0.235 & 0.505 & 123.11 \\
            \midrule
            \texttt{Mistral} & 0.105 & 0.545 & 78.20 & 0.324 & 0.589 & 4.16 & \textbf{0.240} & 0.590 & 25.68 \\
            \texttt{Mistral} w/ $\alpha^*$ & 0.155 & 0.575 & 84.10 & \textbf{0.355} & \textbf{0.649} & \textbf{4.04} & \textbf{0.240} & \textbf{0.640} & 26.60\\
            \texttt{Mistral} w/ 1.2$\alpha^*$ & 0.155 & 0.575 & 88.54 & 0.338 & 0.632 & 4.08 & 0.210 & 0.620 & \textbf{24.87} \\
            \texttt{Mistral} w/ 1.5$\alpha^*$ & \textbf{0.160} & \textbf{0.585} & 88.50 & 0.314 & 0.605 & 4.36 & 0.190 & 0.600 & 29.19\\
            \texttt{Mistral} w/ \texttt{PREADD} & 0.155 & 0.535 & \textbf{49.03} & 0.010$^\text{x}$ & 0.540$^\text{x}$ & 3709.24$^\text{x}$ & \textbf{0.240} & 0.590 & 54.52 \\
            \bottomrule
            \multicolumn{10}{l}{x Invalid results due to unreadable generation containing repeated words or special tokens.}
        \end{tabularx}
\end{table*}

\section{Experiments}
\label{sec:exp}
To show that JAM is capable of scaling to and improving the score on different kinds of tasks, we evaluate our framework following HHH-criteria tasks and the toxicity reduction task. At the same time, we assess JAM with two popular LLMs: \texttt{Meta-Llama-3-8B} \cite{llama3modelcard} and \texttt{Mistral-7B-Instruct-v0.2} \cite{jiang2023mistral7b} to demonstrate its feasibility with various kinds of LLMs. We also compare with the recent work PREADD \cite{pei2023preadd}, which is the incremental work of FUDGE \cite{yang2021fudge}, to show the advantages of JAM. Furthermore, to better illustrate that JAM can help LLMs generate controllable and responsible answers that align with humans, we deploy GPT-4 as a human-like judge. The prompts we used in these experiments can be found at Appendix \ref{appendix:prompt}.

\subsection{HHH Criteria}
We first assess JAM's effectiveness on tasks with different attributes. Using the HHH-criteria \cite{askell2021general} tasks, we show that JAM improves scores across all three criteria, by assessing with multiple metrics provided by the original datasets as listed below:  
\begin{itemize}
    \item Rouge2: a widely used metric to measure the bigram overlap between a generated text and a reference text.
    \item Bleurt: focus on controllability, robustness, and alignment of generated content with reference text. 
    \item Perplexity: a common metric to evaluate the quality of a probabilistic language model.
\end{itemize}

The harmless and helpful datasets are sourced from HH-RLHF \cite{bai2022training}, while the honest dataset is from TruthfulQA \cite{lin2022truthfulqameasuringmodelsmimic}. 


For each criterion, we train a binary classifier using 500 randomly selected data points. Labels for the training data are assigned by T5, assigning a label of 1 if the answer is closer to the correct example and vice versa. We then use the rest of the data points as a test set, ensuring no overlap with the training set to maintain fair evaluation. To thoroughly evaluate our model and align with widely used metrics, we use BIG-bench \cite{srivastava2023beyond} as an evaluation tool. As shown in Table~\ref{table:hhh}, JAM improves scores by up to 22\% with harmless criteria on \texttt{Meta-Llama-3-8B} and consistently outperforms the original models across all criteria.

We also evaluate our method using different moving distances, $\alpha$, as defined in Eq. \ref{eq:cf_mani}, on both models. Here, $\alpha^*$ represents the smallest distance required to shift the decision hyperplane to flip the output label for the desired attribute A = \{\text{Harmless, Honest, Helpful}\}. Then, we test with moving distances $\alpha$ = \{$\alpha^*$, 1.2$\alpha^*$, 1.5$\alpha^*$\}. As expected, increasing $\alpha$ does not yield noticeable improvements. Since our classifier is binary and linear, moving further from the decision boundary does not necessarily reflect an absolute increase in any given attribute. Furthermore, although the scores show substantial improvement, the perplexities of the models with JAM exhibit minimal changes.

The results demonstrate the strong capability of JAM compared to both the baseline models and PREADD. Although PREADD appears to achieve slightly better scores on certain tasks, the results (marked with $^\text{x}$) are deemed invalid as many of the generated outputs are unreadable, often containing repeated words or special tokens. This highlights a significant limitation of PREADD in adapting to different popular models. In comparison, JAM consistently exhibits superior performance across all three tasks.

\begin{table*}[t]
    \centering
	\caption{Comparison of JAM with prompting method using GPT-4 as a judge. The GPT-4 judge is asked to choose the best response in terms of each task's requirement and relevance. Win / Lose / Draw indicates the percentage of times JAM wins, loses, or draws against the prompting method respectively. Here draw is the situation that the GPT-4 judge chooses both.}
	\label{table:hhh_gpt4}
        \begin{tabular}{lccc}
            \toprule
            & Harmless & Honest \\
            \cmidrule(lr){2-2}\cmidrule(lr){3-3}
            & Win / Lose / Draw & Win / Lose / Draw \\
            \midrule
            \texttt{Llama3} w/ $\alpha^*$ & \textbf{0.403} / 0.310 / 0.287 & \textbf{0.500} / 0.238 / 0.262 \\
            \midrule
            \texttt{Mistral} w/ $\alpha^*$ & \textbf{0.162} / 0.158 / 0.680 & \textbf{0.200} / \textbf{0.200} / 0.600 \\
            \bottomrule
        \end{tabular}
\end{table*}

\subsection{GPT-4 Preference}
\label{sec:exp_gpt}
It is also crucial to demonstrate that JAM enables LLMs to generate answers that are not only accurate but also relevant and fluent. To assess this, we compare our method with prompting techniques, which are widely used in CTG. We provide GPT-4 with two answers—one generated by JAM and the other using a prompting approach such as \textit{"You are a \{attribute\} assistant, please..."}. GPT-4 is then asked to choose one of the answers, both (draw), or neither. 




As shown in Table \ref{table:hhh_gpt4}, our method consistently outperforms the prompting approach across all three tasks, achieving a higher winning rate. This further highlights JAM’s ability to control LLMs to generate more coherent answers with desired attributes. For the helpful criterion, however, the ratio of “neither” responses exceeded any chosen answers, making it difficult to determine a clear winner, so we excluded this criterion from our discussion. This result may be related to the inherent capability of the original models in handling this attribute, which limits the potential improvement that could be achieved by the CTG method.

\begin{table*}
    \setlength\tabcolsep{4pt}
    \centering
	\caption{The comparison results on Toxicity Reduction task \cite{gehman2020realtoxicityprompts} with \texttt{Meta-Llama-3-8B} and \texttt{Mistral-7B-Instruct-v0.2} on toxicity score, grammaticality (fluency on grammar level calculated by \texttt{textattack/roberta-base-CoLA}), and perplexity. The results from PREADD marked with~$^\text{x}$~are indicated as invalid because of the generated answers are unreadable and largely affect the Toxicity score.}
	\label{table:tox}
        \begin{tabular}{lccc}
            \toprule
            & \multicolumn{3}{c}{Toxicity Reduction} \\
            \cmidrule(lr){2-4} & {Toxicity}($\downarrow$) & {Grammaticality}($\uparrow$) & {Perpl.}($\downarrow$) \\
            \midrule
            \texttt{Llama3} & 0.122 & \textbf{0.763} & 96.303 \\
            \texttt{Llama3} w/ $\alpha^*$ & \textbf{0.089} & 0.737 & \textbf{94.329} \\
            \texttt{Llama3} w/ \texttt{PREADD} & 0.022$^\text{x}$ & 0.169$^\text{x}$ & 2776.176$^\text{x}$ \\
            \midrule
            \texttt{Mistral} & 0.011 & 0.642 & 135.908 \\
            \texttt{Mistral} w/ $\alpha^*$ & \textbf{0.010} & 0.566 & \textbf{121.358} \\
            \texttt{Mistral} w/ \texttt{PREADD} & 0.066 & \textbf{0.763} & 209.280 \\
            \bottomrule
            \multicolumn{4}{l}{\parbox[t]{9cm}{x Invalid results due to unreadable generation containing repeated words or special tokens.}}
        \end{tabular}
\end{table*}

\subsection{Toxicity Reduction} 
To assess the generalization capability of JAM, we conduct additional evaluations on the toxicity reduction task using the RealToxicityPrompts dataset \cite{gehman2020realtoxicityprompts}, which contains over 100,000 prompts annotated with toxicity scores.

For consistency, we utilize metrics similar to those employed in PREADD. The results are reported for two key metrics, alongside perplexity:
\begin{enumerate}
    \item Toxicity: measure the average toxicity of generated continuations measured via Detoxify \cite{Detoxify}
    \item Grammaticality: measure if structure, syntax, and word order of a sentence conform to the accepted norms of the language by using \texttt{textattack/roberta-base-CoLA} \cite{morris2020textattack})
\end{enumerate}

As shown in Table \ref{table:tox}, JAM demonstrates superior performance in toxicity reduction compared to both the baseline models and PREADD, while maintaining low perplexity and high grammaticality. High grammaticality highlights JAM’s ability to preserve other original attributes of the text, even when steering models toward a specific attribute. In contrast, PREADD with the Llama model still outputs unreadable answers containing repeated words or special tokens and gives low grammaticality and high perplexity.

\subsection{Overhead Analysis}
We also aim to demonstrate the advantage of our method in terms of computational overhead. Due to the simplicity of our classifier, which is a binary and linear model, the training time is negligible. During inference, we record the average computation time for each stage of JAM using randomly selected inputs. Using \texttt{Meta-Llama-3-8B} as an example in Table \ref{table:overhead}, the computational load for generating the manipulation vector is nearly negligible, and the updated generation time is comparable to, or even faster than, the original natural generation. Furthermore, when comparing JAM to the previous CTG method, PREADD, our method demonstrates faster generation speed.

\begin{table*}
\centering
\caption{The average computation times for each stage of JAM using \texttt{Meta-Llama-3-8B}. JAM shows better time efficiency compared with previous CTG work PREADD with the same setting and natural generation. Our method also shows negligible overhead for manipulation operations.}
\label{table:overhead}
\begin{tabular}{llcc}
\toprule
\textbf{Method} & \textbf{Process}& \textbf{Time (seconds)} & \textbf{Time Proportion} \\
\midrule
\multirow{4}{*}{\raisebox{-2ex}{\texttt{JAM}}}
        & Latent Vectors Extraction &1.22 & 54.2\% \\
        \cmidrule(lr){2-4} 
        & Attribute Detection & 0.0006 & 0.03\% \\ \cmidrule(lr){2-4} 
        & Manipulation Vector Generation & 0.0005 & 0.02\% \\ \cmidrule(lr){2-4} 
        & Updated Generation & 1.03 & 45.76\%  \\ \midrule
\texttt{PREADD} & \texttt{PREADD} Generation & 6.51 & -- \\ \midrule
Original & Original Generation & 1.22 & -- \\ \bottomrule
\end{tabular}
\label{tab:comparison}
\end{table*}

\section{Related Works}
Several studies have investigated CTG by modifying a small fraction of model parameters during inference or employing gradient-based methods. For instance, \citet{yang2022unified} adjusts token distributions to prevent over-representation of toxic or biased attributes in the generated text, while \citet{dathathri2020plugplaylanguagemodels} utilizes Plug-and-Play Language Models (PPLM) to steer outputs using classifier gradients. Similarly, FUDGE \cite{yang2021fudge} trains an LSTM-based classifier conditioned on prefix sequences. However, these methods often incur high computational costs due to gradient updates or require extensive training data, which limits their scalability. On the contrary, JAM doesn't require additional training data, demonstrating better scalability. NADO \cite{meng2022controllable} and RepE \cite{zou2023representation} rely on a fine-tuned sequence-to-sequence model and the Low-Rank Representation Adaptation, respectively. These approaches offer approximate solutions that do not consistently meet desired conditions. The lack of transparency and interpretability further diminishes their effectiveness in CTG applications. Inference-Time Intervention \cite{li2024inferencetimeinterventionelicitingtruthful} proposes an approach based on multiple linear classifiers applied to attention heads, but it does not preserve the original latent representation distribution or account for causal relationships among different attributes. Alternatively, JAM directly manipulates the latent space of LLMs to control the generation process, ensuring transparency and interpretability while considering causality.

Alternative frameworks like DATG \cite{liang2024controlled} address CTG via dynamic attribute graphs to control keyword occurrences, but this method involves a time-consuming ranking process, especially for longer contexts, and the need to fine-tune embedding-based classifiers limits scalability across multiple attributes. Similarly, PREADD \cite{pei2023preadd} modifies next-token probabilities by dividing the probabilities generated by a prefix with an undesired attribute from the original probabilities, generating new outputs. Fine-Grained Control via Model Arithmetic \cite{dekoninck2024controlledtextgenerationlanguage} takes a different approach by altering token probabilities through a linear combination of multiple LLMs prompted with different attributes, providing greater flexibility. However, this method is highly sensitive to prompt selection, which can reduce its reliability and scalability. On the other hand, JAM doesn't rely on prompts, being more consistent across different tasks. 

While CAA \cite{panickssery2023steering} follows a different strategy by pre-computing the mean activation difference across numerous pairs of prompts and adding it back during inference, it lacks adaptability to the contextual nuances of user queries, as it relies on a predetermined vector. Conversely, our approach dynamically determines the manipulation distance based on the latent space of the user’s questions, enabling more contextually relevant interventions.

Mix and Match \cite{mireshghallah2022mixmatchlearningfreecontrollable} operates as an iteration-based approach. Likewise, FreeCtrl \cite{feng2024freectrl} modifies feedforward neural network vectors to manage the latent space and continuously re-assesses outputs during the generation process. However, these iterative evaluations can slow down generation, potentially diminishing their practical value. On the contrary, as shown in Table \ref{table:overhead}, JAM only introduces negligible overhead compared to the original generation.

In summary, existing methods either have not effectively and efficiently addressed the challenges or introduced computational overhead while providing interpretable solutions. Those methods are not sufficient to explain and control the generation process of LLMs at a fundamental level.
\section{Conclusion}
In this paper, we proposed a novel framework, JAM, for responsible and controllable text generation. Our framework has demonstrated its strengths in controllable text generation by ensuring both reliability and realism through causality reasoning. It effectively enhances controllable text generation while providing deeper insights with high interpretability through latent vector manipulation. Furthermore, by incorporating a lightweight binary linear classifier with minimal overhead, this explicit and simple design enables seamless adaptation across various attributes and effortless integration with different LLMs. For future work, we are interested in exploring more complex manipulations involving specialized agent models. We are also interested in investigating how controllable text generation can be expanded beyond chatbots and question-answering systems to other applications.


\section{Limitations}
We acknowledge certain limitations of our method. When handling more complex requests, such as implicit preferences, the decision boundary may be insufficient in a single linear space. One potential approach to mitigate this issue is through the linear combination of decision boundaries, which requires further investigation. This includes exploring implicit attributes and developing more efficient and effective methods for latent space manipulation. Additionally, while JAM is computationally efficient, it sometimes requires LLMs to generate twice for attribute detection and effective latent space manipulation. Due to time and scope limitations, we have not addressed these issues in the current work, but we plan to tackle these limitations in the future.

\section{Acknowledgment}
We acknowledge the use of AI assistants based on GPT-4 to help us with the preparation of this manuscript.
\newpage
\bibliography{reference}
\appendix
\newpage
\section{Prompt Information}
\label{appendix:prompt}
In this section, we provide the prompt information in each of our experiments:
\begin{itemize}
    \item The prompts for all \textbf{HHH-criteria} experiments are as follows:
        \begin{quote}
        \textit{"Human asked you a question, please answer as an assistant. \textcolor{orange}{\{Question\}}"}
        \end{quote}
    \item The prompts for all \textbf{Toxicity Reduction} experiments are as follows:
        \begin{quote}
        \textit{"\textcolor{orange}{\{Prompt\}}. Given the prompt, continue the sentences."}
        \end{quote}
    \item The prompts for \textbf{GPT-4} are as follows:
        \begin{quote}
        \textit{"The user did an experiment where several language models were prompted to continue the conversation with an input message written by a person. The responses of the language models are:} \textit{Person 1: \textcolor{orange}{\{Response 1\}}} \textit{Person 2: \textcolor{orange}{\{Response 2\}}}
        
        \textit{This input message often contains \textcolor{orange}{\{Attribute\}} language and the goal is to continue the conversation with a message that is \textcolor{orange}{\{Desired Attribute\}}. ... Briefly reason about which continuation is better and end your answer with: "Preferred continuation: 1" or "Preferred continuation: 2", depending on which option is better. If both options are equally good, end your response with "Preferred continuation: None". If both options are equally bad, end your response with "Preferred continuation: Neither"."}
        \end{quote}
\end{itemize}

\end{document}